\documentclass[letterpaper, 10 pt, conference]{ieeeconf}  

\IEEEoverridecommandlockouts                              

\usepackage{multirow, makecell}
\usepackage{hhline}
\usepackage{array, booktabs}

\usepackage{graphicx}
\usepackage{enumerate}
\usepackage{epsfig} 
\usepackage{subcaption}
\usepackage{amsmath} 
\usepackage{amssymb}  

\usepackage{bm}
\usepackage{multirow, makecell}
\usepackage[linesnumbered,ruled,vlined]{algorithm2e}
\usepackage{cite}
\usepackage{hyperref}
\usepackage{cleveref}
\usepackage{algpseudocode}
\usepackage[font={small}]{caption}
\usepackage{color}

\title{\LARGE \bf
Generic Prediction Architecture Considering both Rational and Irrational Driving Behaviors
}

\author{Yeping Hu, Liting Sun and Masayoshi Tomizuka 
\thanks{*This work was supported by Momenta.}
	\thanks{Y. Hu, L. Sun and M. Tomizuka are with the Department of Mechanical Engineering, University of California, Berkeley, CA 94720 USA {\tt {[yeping\_hu, litingsun, tomizuka@berkeley.edu]}}}
}

\makeatletter
\def\endthebibliography{%
	\def\@noitemerr{\@latex@warning{Empty `thebibliography' environment}}%
	\endlist
}
\makeatother
\begin{document}
\bstctlcite{IEEEexample:BSTcontrol}

\maketitle
\thispagestyle{empty}
\pagestyle{empty}

\begin{abstract}
Accurately predicting future behaviors of
surrounding vehicles is an essential capability for autonomous vehicles in order to plan safe and feasible trajectories. The behaviors of others, however, are full of uncertainties. Both rational and irrational behaviors exist, and the autonomous vehicles need to be aware of this in their prediction module. The prediction module is also expected to generate reasonable results in the presence of unseen and corner scenarios. Two types of prediction models are typically used to solve the prediction problem: learning-based model and planning-based model. Learning-based model utilizes real driving data to model the human behaviors. Depending on the structure of the data, learning-based models can predict both rational and irrational behaviors. But the balance between them cannot be customized, which creates challenges in generalizing the prediction results. Planning-based model, on the other hand, usually assumes human as a rational agent, i.e., it anticipates only rational behavior of human drivers. In this paper, a generic prediction architecture is proposed to address various rationalities in human behavior. We leverage the advantages from both learning-based and planning-based prediction models. The proposed approach is able to predict continuous trajectories that well-reflect possible future situations of other drivers. Moreover, the prediction performance remains stable under various unseen driving scenarios. A case study under a real-world roundabout scenario is provided to demonstrate the performance and capability of the proposed prediction architecture.

\end{abstract}

\section{Introduction}
\subsection{Motivation}
While interacting with human drivers, autonomous vehicles need to be aware of the multi-modal interaction outcomes in order to have reasonable prediction results. Such multimodality comes from the fact that human can have different levels of rationality. Rational behaviors usually result in safe and feasible driving motions, while irrational behaviors are typically reflected by dangerous and unusual driving motions that may lead to car accidents. 

Even though most people would consider themselves to be logical and can make rational decisions, drivers are not necessarily absolute rational. They do not always drive in optimal and safe trajectories since sometimes they are willing to take risks for their own benefits. For example, a driver may perform a dangerous cut-in maneuver to change into a desired lane quickly; a novice driver might be overly cautious and tends to make improper braking. Besides, drivers can easily have irrational driving behaviors due to wrong predictions of other road entities or unawareness of the surroundings. Therefore, autonomous cars are expected to consider not only rational but also irrational behaviors of other drivers during the prediction process, which will assure preparation for potential emergencies as well as safe and comfortable driving experiences.



\subsection{Related Works}
\subsubsection{Learning-based Approach}
Learning-based methods \cite{traj_LSTM1, HMM_GP, traj_LSTM3, Yeping_MDN, traj_DBN1, traj_HMM1, traj_image1} have been wildly used for prediction problems for autonomous vehicles, which utilize real data to produce the future outcomes of human drivers. In \cite{HMM_GP}, the authors modeled the driver behavior by hidden Markov models (HMM) and Gaussian Process (GP) to generate a group of future trajectories of the predicted vehicle. The long short-term memory (LSTM) method is utilized in \cite{traj_LSTM1} and \cite{traj_LSTM3}  to analyze past trajectory data and predict the future locations of the surrounding vehicles. \cite{Yeping_CVAE} proposed to combine a modified mixture density network (MDN) \cite{Yeping_MDN} and a conditional variational autoencoder (CVAE) to predict both discrete intention and continuous motions for multiple interacting vehicles. Based on generative models, an interpretable multi-modal prediction method is proposed in \cite{Yeping_2019IV} , which can predict interactive behavior for traffic participants.

The main advantage of learning-based methods is that complicated models can be learned to represent real-life situations for prediction. However, two drawbacks need to be concerned for the learning-based method. First is the data insufficiency. Although researchers have tried to use more training data to learn driving models, it is nearly impossible to have a dataset that is large enough to cover every possible driving situations and thus the learned model can easily fail under any unseen or corner cases. Second is the inherent biases of the collected data. In fact, if the training data has some inherent biases, the driving model will not only learn those biases but will end up amplifying them. For example, if under a certain driving scenario, the data contain mostly irrational behaviors, the learned model will be inclined to predict more irrational than rational behaviors. 

\subsubsection{Planning-based Approach}

Planning-based approaches \cite{A_star1, A_star2, A_star3, RRT_1, RRT_2, RRT_3} assume that human drivers are approximately optimal planners with respect to some reward functions, i.e., their future trajectories are maximizing their rewards. Hence, the prediction of their trajectories can be obtained by solving optimization problems with the correct reward functions. 
To acquire such reward functions, inverse reinforcement learning (IRL) has been widely adopted. It aims at finding a reward function which can match best in terms of key features with the observed demonstrations. Initially proposed by Kalman \cite{kalman}, \cite{ng2000algorithms} and \cite{abbeel2004apprenticeship} formulated it as apprenticeship learning by maximizing the margin in terms of feature matching. Later, Ziebart \textit{et al.} \cite{ziebart2008maximum} further extended IRL to deal with the probabilistic characteristic of reward functions via the maximum entropy principle. Levine \textit{et al.} \cite{Levine2012ICML} proposed to solve the maximum-entropy IRL directly in continuous domain and applied it to predict human driving behaviors. Based on these works, many IRL based prediction or human behavior modeling approaches have been proposed. For instance, \cite{kretzschmar2016socially} learned a model for cooperative agents to generate human navigation behavior, and in \cite{sun2018courteous}, the authors used IRL to model the social impact between interactive agents. In \cite{Liting_IRL}, the authors formulated a hierarchical IRL to model human driver's decision making and trajectory planning, so that interactive driving behaviors can be predicted.  


One main advantage of the aforementioned planning-based methods is their inherent implication of causality. Hence, they can easily guarantee the feasibility of the predicted trajectories and have better generalization ability to unseen circumstances. For instance, \cite{Sun2019IV} used planning-based methods to infer about uncertainties for better prediction. However, the planning-based approach also suffer from several issues. First, most of the planning-based approaches assume that all drivers are approximately rational, i.e., they are optimizing some reward/cost function while driving. This assumption cannot be strictly hold in practice since ``irrational'' behaviors are inevitable. Consequently, the predicted trajectories might be too conservative to reflect potential dangers in some situations. Although recent works such as \cite{majumdar2017risk} and \cite{Sun2019ITSC} have introduced some ``irrational'' behaviors into planning-based approaches, it is still not sufficient to cover all human driving behaviors. Moreover, to obtain a solvable and interpretable planning problem, the learned reward/cost functions are typically linear combinations of features. Such representation might not be complicated enough to capture different driving behavior.



\subsection{Contribution}
As planning is a lower module of prediction, it is common to solve planning problems by utilizing prediction results to incorporate uncertainties. However, very few works explicitly apply the planning approach in the prediction problem. In this work, we propose to leverage the advantages of both the planning-based and learning-based methods to mutually compensate for their drawbacks, which can enhance the overall prediction performance. The main contributions of this work are as follows: 
\begin{itemize}
	\item We propose a generic environmental representation methodology for vehicle behavior predictions under highly-interactive scenarios. 
	\item Both the irrational and rational future trajectories of human drivers can be predicted, which can provide better environment information to the autonomous vehicle.
	\item The performance of the proposed architecture remains stable and can be safely used under rare events and corner cases.
	\item A challenging real-world roundabout scenario is considered in this work to demonstrate the capability of our method.
\end{itemize}

\section{Problem Statement}


\subsection{A mathematical representation} 
In this problem, we consider the interactive behavior between two vehicles: the ego autonomous vehicle (denoted by $(\cdot)^{ego}$) and the predicted human-driven vehicle (denoted by $(\cdot)^{pred}$). Other vehicles in the scene will be regarded as surrounding vehicles which is denoted by $(\cdot)^{surr}$. For a given vehicle, we use $\xi$ to represent historical trajectories, $\hat{\xi}$ for future trajectories, and $\hat{\xi}_{gt}$ for ground truth future trajectories. We store all historical trajectories of the related vehicles in $\xi$, where  $\xi = [\xi^{pred}, \xi^{ego}, \xi^{surr}]$. Note that a trajectory is represented by a sequence of states of the vehicle, i.e. $\xi = [x_1, x_2, \dots, x_T]$ where $T$ is the trajectory horizon and $x_t$ denotes the vehicle state at $t$-th time step.

We aim at predicting the behavior of the selected vehicle while considering the potential influence of its future behavior from our own vehicle. We use the following conditional probability density function (PDF) to represent the correlated future trajectories of two interactive vehicles:
\begin{eqnarray}
P(\hat{\xi}^{pred}, \hat{\xi}^{ego} | \xi),
\end{eqnarray}
where the possible joint trajectories of the two vehicles depend on the historical trajectories of their own as well as all surrounding vehicles in the scene. However, after the ego vehicle planned a possible future trajectory for its own, it is necessary to marginalize out $\hat{\xi}^{ego}$ to obtain the future trajectories of the predicted vehicle conditioned on the ego vehicle's planned trajectory. Mathematically, such dependencies can be expressed as:
\begin{eqnarray}
P(\hat{\xi}^{pred} |  \hat{\xi}^{ego}, \xi).
\end{eqnarray}

In general, we are trying to infer possible future behaviors of the predicted vehicle given several potential future trajectories planned for the ego vehicle. 

\subsection{Generic Environmental Representation}
\begin{figure}[htbp]
	\centering
	\includegraphics[scale=0.8]{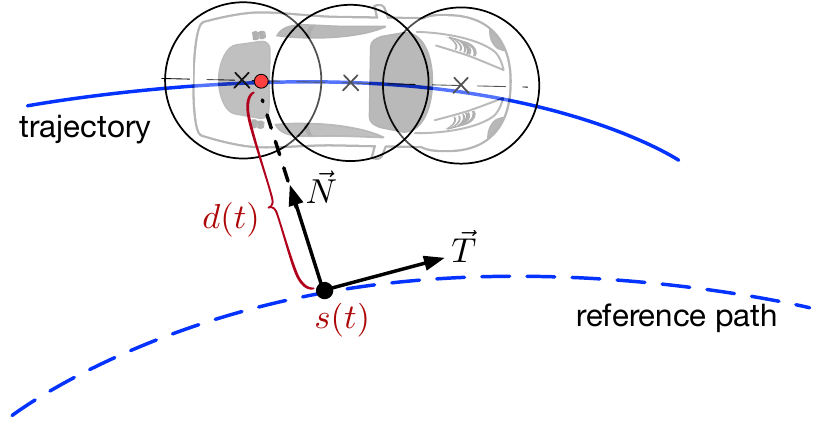}
	\caption{Illustration of the Fren\'{e}t frame and vehicle boundaries. The red point represents the vehicle's center of mass. The shape of the vehicle is approximated by three circles.}
	\label{fig:frenet}
\end{figure}
\subsubsection{Representation in Fren\'{e}t Frame}
Instead of Cartesian coordinate, we utilized the Fren\'{e}t Frame to represent vehicle state. 
As illustrated in Fig.~\ref{fig:frenet}, the vehicle motion in the Fren\'{e}t Frame can be represented with the longitudinal position along the path $s(t)$, and lateral deviation to the path $d(t)$. Therefore the vehicle state at time step $t$ can be defined as $x_t = (s(t), d(t))$. Note that the reference path of a vehicle will change according to the road it is current driving on. The origin of the reference path can be defined differently according to different objectives and each reference path will have its own Fren\'{e}t Frame. Since we are dealing with interaction between two vehicles, we define the origin as the cross point of their reference path. Note that each vehicle can have several possible reference paths, thus there could be multiple combinations of reference path pair corresponding to different cross points. 
\subsubsection{Conversion between Cartesian coordinate and Fren\'{e}t Frame}
For both ego and predicted vehicle, the following steps need to be performed throughout the testing process of the proposed architecture:

\begin{itemize}
    \item [(i)] Determine the possibility of every reference trajectory according to historical path using dynamic time wrapping (DTW) \cite{DTW_origin}\cite{Yeping_2019IV}.
    \item [(ii)]  Map the current global position (in Cartesian coordinate) of the vehicle on to each possible reference path (in Fren\'{e}t Frame).
    \item [(iii)] Perform the proposed prediction algorithm under the Fren\'{e}t Frame.
    \item [(iv)] Convert the predicted results back to Cartesian coordinate to check for collision and visualize the result.
    \item [(v)] Receive a new observation in Cartesian coordinate and repeat step (i)-(iv). 
\end{itemize}



\begin{figure*}
	\centering
	\includegraphics[scale=0.6]{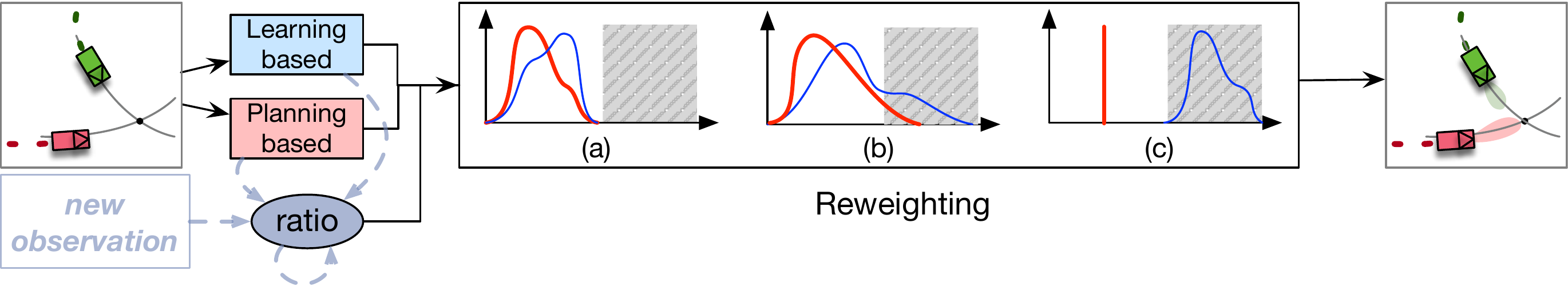}
	\caption{This plot shows the process of proposed architecture. Note that trajectories directly sampled from the learning-based method can have three different degrees of rationality: (a) fully rational; (b) partially rational; (c) fully irrational. Here, we demonstrate intuitive illustration of the reweighting process in each of these cases, where the original probability density function (PDF) of all sampled trajectories is shown in blue, the PDF after the reweighting process is shown in red, and the shaded area contains infeasible sample points.}
	\label{fig:architecture}
\end{figure*}
\section{Prediction Architecture}
\label{sec: prediction_arch}
In this section, we first introduce the detailed formulation and structure of the overall prediction framework. We then overview the approaches used for the learning-based and planning-based model. The graph illustration of the proposed architecture as well as the pseudo-code of the algorithm are shown in Fig.~\ref{fig:architecture} and Algorithm 1 respectively.  

\subsection{Overall Framework}
\label{subsec: framework}
The proposed generic prediction framework contains the following six steps:
\subsubsection{Sample Joint Trajectories}
Given the observed historical trajectories of the two interacting vehicles and their surrounding vehicles, the future joint trajectories can be sampled from the predicted joint distribution generated by a learning-based method. The $N$ sampled trajectory pairs can be expressed as $\{ \hat{\xi}^{pred}, \hat{\xi}^{ego}\}_{1:N}$.

\subsubsection{Convert to Conditional Distribution}
Since we are interested in predicting the possible future trajectories of the predicted vehicle given the most likely future trajectory of the ego vehicle, we need to convert the predicted joint distribution into the a conditional distribution. Namely, we want to get $P(\hat{\xi}^{pred}|\hat{\xi}_{gt}^{ego}, \xi)$ from $P(\hat{\xi}^{pred}, \hat{\xi}^{ego}|\xi)$. According to Bayes rule, we have:
\begin{eqnarray}
\label{eqn:1}
P(\hat{\xi}^{pred}|\hat{\xi}_{gt}^{ego}, \xi) = \frac{P(\hat{\xi}^{pred}, \hat{\xi}_{gt}^{ego}|\xi)}{\sum_{\Tilde{\xi}^{pred}}P(\Tilde{\xi}^{pred}, \hat{\xi}_{gt}^{ego}|\xi)},
\end{eqnarray}
where the numerator is an unknown distribution since it is nearly impossible to obtain a sampled future trajectory of the ego vehicle that exactly equals to the ground-truth. To resolve this problem, we first rewrite $\hat{\xi}^{ego}$ as $\hat{\xi}_{gt}^{ego} + \Delta\xi^{ego}$, where $\hat{\xi}_{gt}^{ego}$ denotes the ground truth future trajectory of the ego vehicle and $\Delta\xi^{ego}$ represents the discrepancy between the ego vehicle's ground truth and each of its possible future trajectory. Then among all the sampled joint trajectories, if any $\Delta\xi^{ego}$ is smaller than a certain threshold, we denote the predicted ego vehicle's trajectory as a satisfied trajectory $\hat{\xi}_{s}^{ego}$ that can well approximate $\hat{\xi}_{gt}^{ego}$ and we store the corresponding trajectory pair $\{ \hat{\xi}^{pred}_{s}, \hat{\xi}_{s}^{ego}\}$. Finally, for all saved samples, we can rewrite (\ref{eqn:1}) as:
\begin{equation}
\begin{split}
P(\hat{\xi}^{pred}_{s}|\hat{\xi}_{gt}^{ego}, \xi) &\approx P(\hat{\xi}^{pred}_{s}|\hat{\xi}_{s}^{ego}, \xi) \\ & = \frac{P(\hat{\xi}^{pred}_{s}, \hat{\xi}_{s}^{ego}|\xi)}{\sum_{\Tilde{\xi}^{pred}_{s}}P(\Tilde{\xi}^{pred}_{s}, \hat{\xi}_{s}^{ego}|\xi)} \\ & \approx P(\hat{\xi}^{pred}_{s}, \hat{\xi}_{s}^{ego}|\xi).
\end{split}
\end{equation}

Note that the ground-truth trajectory of the ego vehicle is unknown if the algorithm is running online. Therefore, the ego vehicle needs to first plan for itself in real-time and use each of the planned trajectories as its ground-truth trajectory. 


\subsubsection{Optimal Trajectory Generation}
Apart from sampling trajectories that may contain both rational and irrational behaviors, we can further generate the most probable trajectory by solving a finite horizon Model Predictive Control (MPC) problem using the learned continuous cost function via a planning-based method, which guarantees the resulting behavior be rational. The equation of obtaining the optimal trajectory can be expressed as:
\begin{eqnarray}
\hat{\xi}_{opt}^{pred} = \arg\min_{\Tilde{\xi}^{pred}}C(\bm{\theta}, \Tilde{\xi}^{pred}, \hat{\xi}_{gt}^{ego}, \xi),
\end{eqnarray}
where $\bm{\theta}$ is the weight parameter vector for cost function $C$ to determine the importance of each of the features.

\subsubsection{Weight Ratio Update}
The next step is to determine how many optimal trajectory pairs $\hat{\xi}^{pred}_{opt}$ should we add to the trajectory set so that the resampling result are reasonable. In fact, we want to keep updating the probability of both the optimal trajectory $P_{opt}$ and the average of the satisfied trajectories obtained from the learning-based prediction approach $P_{s}$. Note that $P_{opt} + P_{s}  = 1$  and we denote the ratio of the two probabilities as $r$ where $r = \frac{P_{s}}{P_{opt}}$. The probabilities are calculated by comparing the similarities between the actual observed trajectory and the predicted trajectory for the predicted vehicle, which will be updated after each new observation at the next time step using Bayes' theorem. The number of added optimal trajectory pairs $N_{opt}$ is then equal to $r \times N_s$, where $N_s$ denotes the total number of satisfied trajectories $\hat{\xi}^{pred}_{s}$ from step 2). Therefore, the ratio $r$ can be also regarded as the weight ratio between the optimal and the satisfied trajectory pairs. 


\subsubsection{Distribution Reweighting}

After having all the satisfied trajectories sampled from the learning-based method along with the added optimal trajectories from the planning-based method, we are then able to re-evaluate each trajectory's probabilities using our learned cost function. According to the principle of maximum entropy, the distribution of agents' behaviors can be approximated by an exponential distribution family and thus we can write
\begin{eqnarray}
P(\hat{\xi}^{pred}|\hat{\xi}_{gt}^{ego}, \xi) \propto \exp^{-C(\bm{\theta},\hat{\xi}^{pred}, \hat{\xi}_{gt}^{ego}, \xi)},
\end{eqnarray}
where we will minimize the probability of irrational behaviors that might generate infeasible trajectories while increasing the likelihood of rational behaviors that result in safe trajectories. 

\subsubsection{Resampling}
Finally, we need to resample trajectories from the sample set according to the updated conditional distribution obtained from the previous step. These sampled trajectories are then regarded as our final prediction results at the current time step. 

\SetKwInOut{TrainedModel}{Trained Models}
\SetKwInOut{Input}{Input}
\SetKwInOut{Output}{Output}
\SetKwInOut{SensorInput}{Sensor}
\SetKwInOut{Return}{Return Prediction}

\begin{algorithm}
    \LinesNumberedHidden
	\caption{Proposed Generic Prediction Architecture}
	\label{alg:hybrid system}
	\SetAlgoLined
	\DontPrintSemicolon
	\Input{$\xi$ , map information}  
	\Output{$\hat{\xi}^{pred}$ at each prediction time step} 
	\TrainedModel{$\mathcal{M}_L$ - learning-based model \\  $\mathcal{M}_P$ - planning-based model}

	\While {algorithm is running}{
		1) $\{\hat{\xi}^{pred}, \hat{\xi}^{ego}\}_{1:N}$$\leftarrow$sample joint trajectories ($\mathcal{M}_L$)\\ 
		2) $\hat{\xi}^{ego}_{gt}$ $\leftarrow$ path planning for ego vehicle\\  \quad  $\{\hat{\xi}^{pred}, \hat{\xi}^{ego}_{s}\}_{1:N_s}$$\leftarrow$filter trajectories and update sample set\\ 
		3) $\hat{\xi}^{pred}_{opt}$$\leftarrow$generate optimal trajectory ($\mathcal{M}_P$ + MPC)\\ 
		4) $r$ $\leftarrow$ update weight ratio\\  \quad
		$N_{opt} = r \times N_s$ $\leftarrow$ add $N_{opt}$ optimal trajectories to the current sample set\\ 
		5) $P(\hat{\xi}^{pred}|\hat{\xi}^{ego}_{gt}, \xi, z)$ $\leftarrow$ reweight samples ($\mathcal{M}_P$) \\  
		6) $\hat{\xi}^{pred}$ $\leftarrow$ resample from the updated conditional distribution \\  
		\Return{$\hat{\xi}^{pred}$}
		Obtain new state observation from the sensor. \\  
		$\xi$, map information $\leftarrow$ update
	    }
\end{algorithm}

\subsection{Learning-based Trajectory Prediction}
The learning-based trajectory prediction method we applies is called the conditional variational autoencoder (CVAE) \cite{Yeping_2019IV}\cite{CVAE_1}, which is a latent variable model that is rooted in Bayesian inference. The goal is to model the underlying probability distribution of the data using a factored, low-dimensional representation. In this problem, our objective is to estimate the probability distribution of joint trajectories in (1) by utilizing the encoder-decoder structure of CVAE. 

The encoder $\mathcal{Q}_\varphi$, parameterized by $\varphi$, takes the input $X$ as a learned embedded space of historical trajectories of all vehicles ($\xi$) and $Y$ as the actual future trajectories of the two interacting vehicles ($\hat{\xi}^{ego}_{gt}, \hat{\xi}^{pred}_{gt}$) to ``encode" them into a latent $z$-space. Then the decoder $\mathcal{P}_\psi$, parameterized by $\psi$, takes $X$ and sampled $z$ values from the latent space to ``decode" them back to the future trajectories $\hat{Y}$ as the prediction result ($\hat{\xi}^{ego}, \hat{\xi}^{pred}$). The network is trained using the reparameterization trick \cite{VAE_original} to enable back-propagation, where the network tries to minimize the evidence lower bound (ELBO) and it is formulated as:
\begin{equation}
\begin{split}
	\mathcal{L} = -\mathbb{E}_{\mathcal{Q}_\varphi}&\big[\log \mathcal{P}_\psi(Y|X, z)\big] + \beta D_{KL}(\mathcal{Q}_\varphi(z|X, Y)||p(z)),
\end{split}
\end{equation}
where $p(z)$ denotes the prior distribution of the latent $z$ space and it is usually defined as a unit Gaussian. The overall idea of the loss function is to have a good estimation of data log-likelihood as well as a small Kullback-Leibler (KL) divergence denoted by $D_{KL}$
between the approximated posterior and the prior $p(z)$ at the same time. The hyperparameter $\beta$ is used to control the training balance between the two losses for better performance.

Note that during the test time, only the decoder will be used. Each predicted joint trajectories will be generated when we randomly sample one $z$ value and feed it into the decoder network along with the historical input $X$.

\subsection{Planning-based Trajectory Prediction}
The planning-based trajectory prediction stems from Theory of Mind \cite{premack1978does} which describes the prediction process of human. We let the ego vehicle simulate what the other vehicle will do assuming that it is approximately optimal planners with respect to some reward or cost functions, i.e., it is a noisily rational driver. Under this assumption, the target vehicle's driving behavior can be described via its cost function which can be learned based on demonstrations. In this paper, we adopt the continuous domain maximum-entropy inverse reinforcement learning (IRL) \cite{ziebart2008maximum, Levine2012ICML}. A brief review of the algorithm is given below.

Assume that the cumulative cost $C$ of the target vehicle is a linear combination of a set of selected features over a defined horizon $N$. Then given a demonstration set $U_D$ which contains $M$ interactive trajectories of both the ego and target vehicles:
\begin{IEEEeqnarray}{rCl}
U_D = \{(\xi^{pred}_{gt,i}, \xi^{ego}_{gt,i}, \xi), i=1,2,\cdots,M\}, 
\end{IEEEeqnarray}
we can write the cumulative cost function as
	\begin{equation}
		C(\xi^{pred}_{gt}{,}\xi^{ego}_{gt}{,}\xi{;}\theta)=\theta^T \sum_{t=0}^{N{-}1}\mathbf{\phi}(x^{pred}_t{,} x^{ego}_t{,}\xi),
		\label{eq:selfish_cost_cumulative}
	\end{equation}
where $\mathbf{\phi}$ is the feature vector which includes the distance between two vehicles, the speed gap with respect to the speed limit, the acceleration, and the lateral deviation from the target lane. $x^{pred}_t$ and $x^{ego}_t$ are, respectively, the states of the target vehicle and the ego vehicle at time instant $t$ within the planning horizon.

Building on the principle of maximum entropy, we assume that trajectories are exponentially more likely when they have lower cost:
	\begin{equation}
	P(\xi^{pred}_{gt}|\xi^{ego}_{gt}, \xi) \propto \exp\left(-C(\xi^{pred}_{gt}{,}\xi^{ego}_{gt}{,}\xi{;}\theta)\right).
	\end{equation}
Then, our goal is to find the weight $\theta$ which maximizes the likelihood of the demonstration set ${U}_{D}$:
	\begin{IEEEeqnarray}{rCl}
	\theta^*&=&\arg\max_{\theta}P({U}_{D}|\theta)\\
	&=&\arg\max_{\theta}\Pi_{i=1}^{M}\dfrac{ P(\xi^{pred}_{gt,i}|\xi^{ego}_{gt,i}, \xi_i, \theta) } {P(\theta)}\\
	&=&\arg\max_{\theta}\Pi_{i=1}^{M}\dfrac{ P(\xi^{pred}_{gt,i}|\xi^{ego}_{gt,i}, \xi_i, \theta) }{\int P(\tilde{\xi}^{pred}_{gt}|\xi^{ego}_{gt,i}, \xi_i, \theta) d\tilde{\xi}^{pred}_{gt}}
	\label{eq:optimal_lambda}
	\end{IEEEeqnarray}
	
	To tackle the partition term ${\int P(\tilde{\xi}^{pred}_{gt}|\xi^{ego}_{gt,i}, \xi_i, \theta) d\tilde{\xi}^{pred}_{gt}}$ in (\ref{eq:optimal_lambda}), we approximate $C$ with its Laplace approximation as proposed in \cite{Levine2012ICML}:
	\begin{IEEEeqnarray}{rCl}
		C(\tilde{\xi}^{pred}_{gt}{,}\xi^{ego}_{gt,i}{,}\xi_i{;}\theta)&\approx& C({\xi}^{pred}_{gt,i}{,}\xi^{ego}_{gt,i}{,}\xi_i{;}\theta)\nonumber\\
		&&{+}\left(\tilde{\xi}^{pred}_{gt}{-}{\xi}^{pred}_{gt,i}\right)^{T}\dfrac{\partial C}{\partial {\xi}^{pred}_{gt}}\nonumber\\
		&&+\dfrac{1}{2}\left(\tilde{\xi}^{pred}_{gt}{-}{\xi}^{pred}_{gt,i}\right)^T\dfrac{\partial^2 C}{{\partial} {\xi}^{pred}_{gt}}\times\nonumber\\
		&&\qquad\left(\tilde{\xi}^{pred}_{gt}{-}{\xi}^{pred}_{gt,i}\right).
		\label{eq:laplace_approximation}
	\end{IEEEeqnarray}
	
	\begin{figure}[htbp]
	\centering
	\includegraphics[scale=0.33]{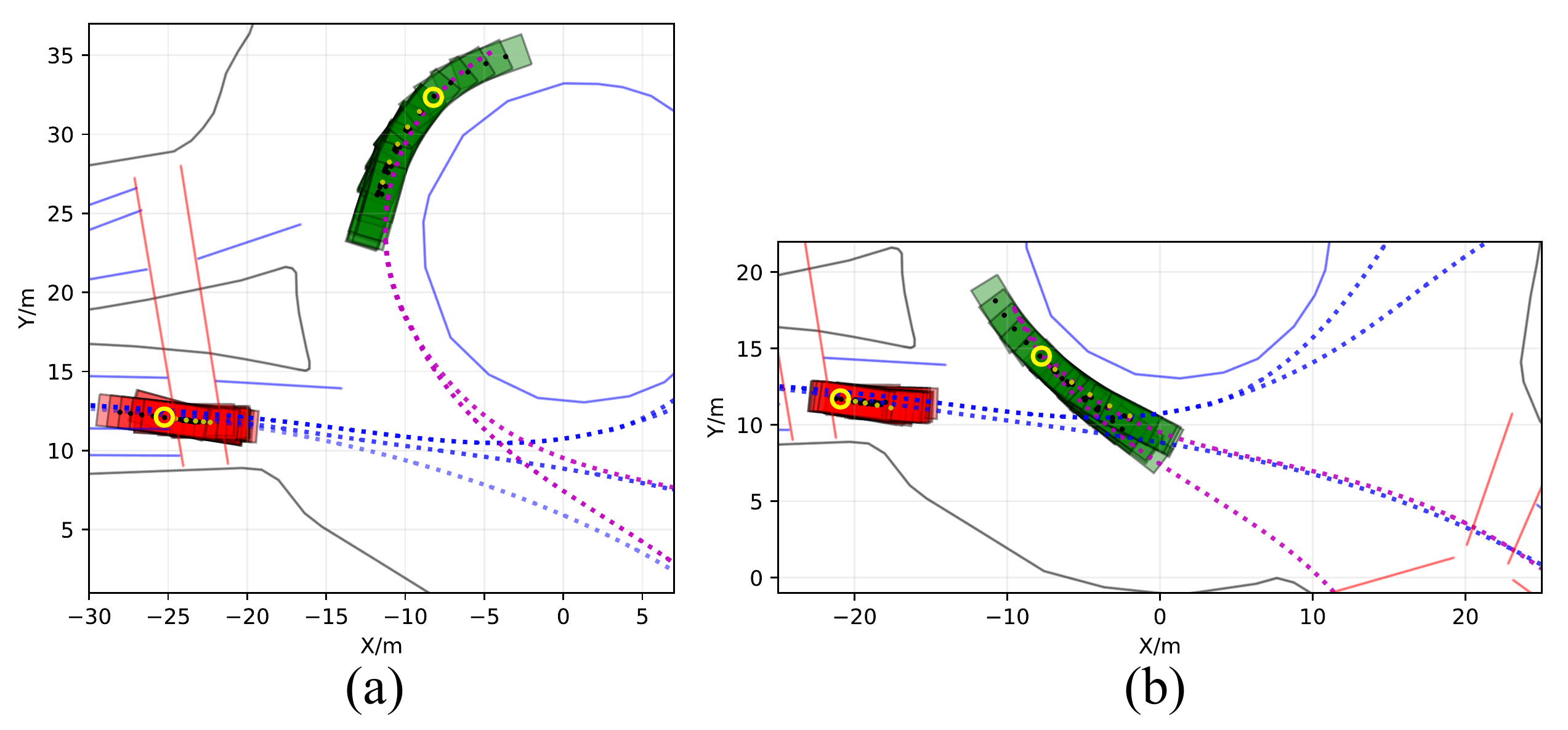}
	\caption{Visualized prediction results of a selected scene at different time steps. The red car is the ego vehicle and the green car is the predicted vehicle. We plotted both the historical and the predicted trajectories of each vehicle, where the yellow circles represent the current state of two vehicles. The small yellow dots denote ground-truth states and the dashed lines are the possible reference paths for both vehicles.}
	\label{fig:multi_route}
\end{figure}
\begin{figure}[htbp]
	\centering
	\includegraphics[scale=0.33]{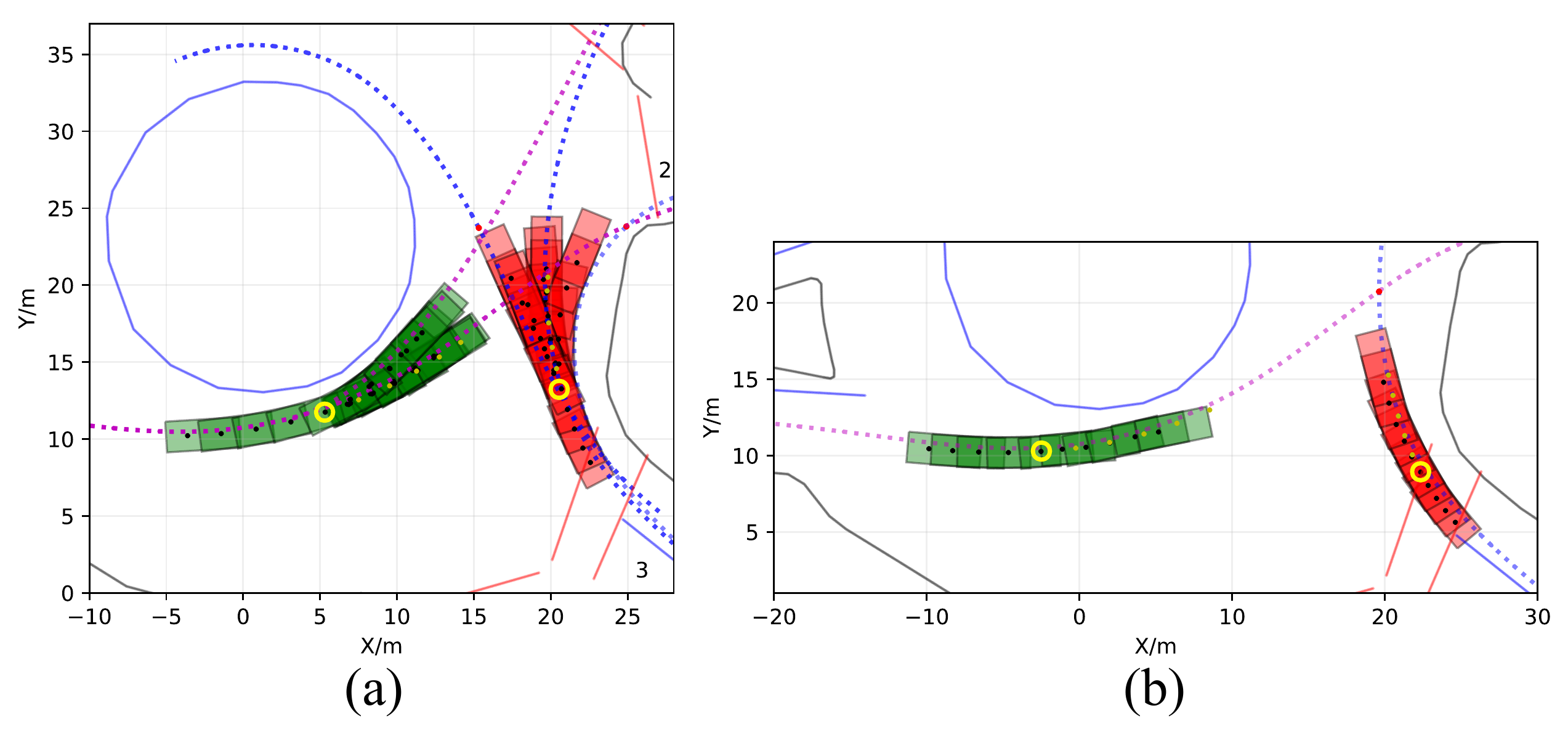}
	\caption{Generalization results at an unseen roundabout entrance.}
	\label{fig:generalization}
\end{figure}
	With the assumption of locally optimal demonstrations, we have $\dfrac{\partial C}{\partial {\xi}^{pred}_{gt}}|_{{\xi}^{pred}_{gt,i}}{\approx}0$ in (\ref{eq:laplace_approximation}). This simplifies the partition term ${\int P(\tilde{\xi}^{pred}_{gt}|\xi^{ego}_{gt,i}, \xi_i, \theta) d\tilde{\xi}^{pred}_{gt}}$ as a Gaussian Integral where a closed-form solution exists (see \cite{Levine2012ICML} for details). Substituting (\ref{eq:laplace_approximation}) into (\ref{eq:optimal_lambda}) yields the optimal parameter $\theta^*$ as the maximizer.
	
	Once we have learned the cost function, we can use it to either evaluate the probabilities of given trajectory samples, or generate the most probable trajectory by solving a MPC problem, as explained in Section \ref{subsec: framework}.

%

\begin{figure*}
\centering
\begin{minipage}{.65\textwidth}
  \centering
  \includegraphics[scale=0.22]{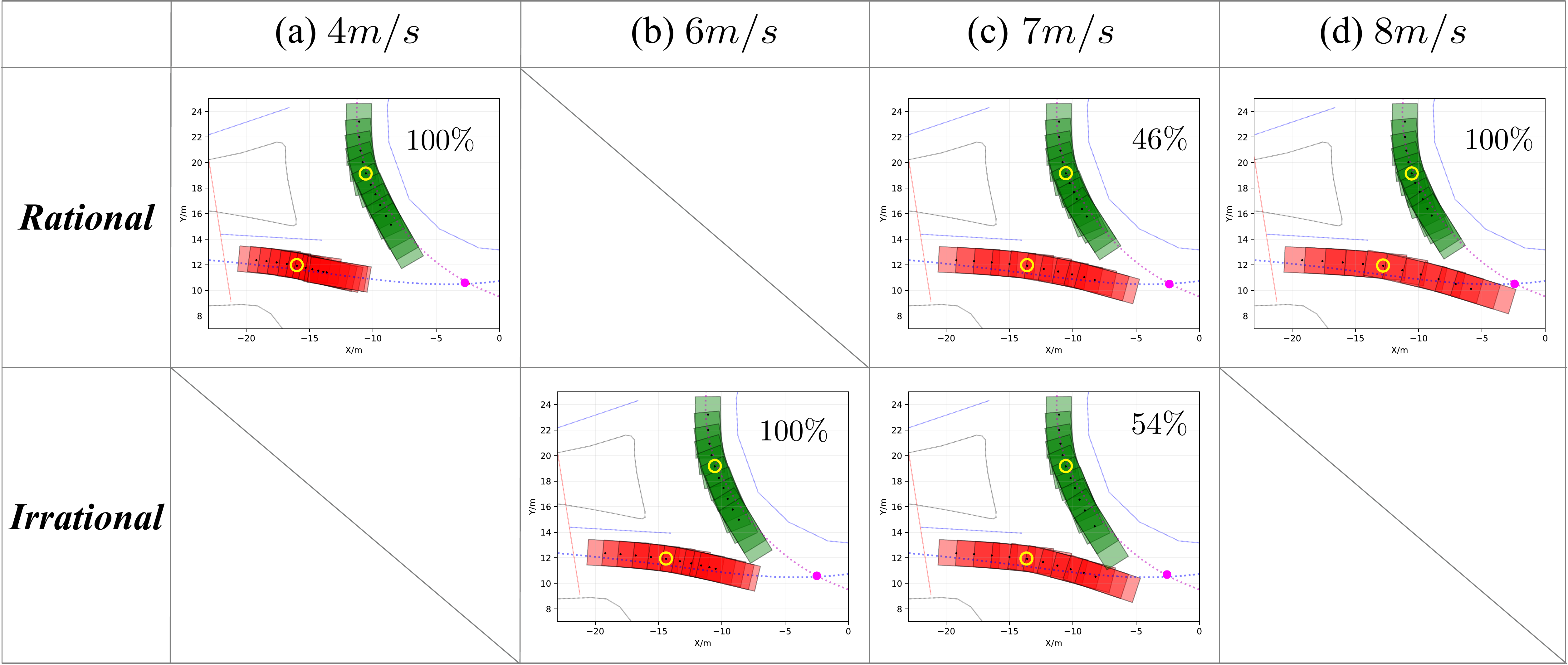} 
  \caption{Selected artificial test scenarios and the corresponding results. The pink dot represents the cross point of two vehicles’ ground-truth reference paths. The number on the top right corner denotes the percentage rate of rational or irrational behaviors.}
  \label{fig:artificial_test}
\end{minipage}%
\begin{minipage}{.35\textwidth}
  \centering
  \includegraphics[scale=0.43]{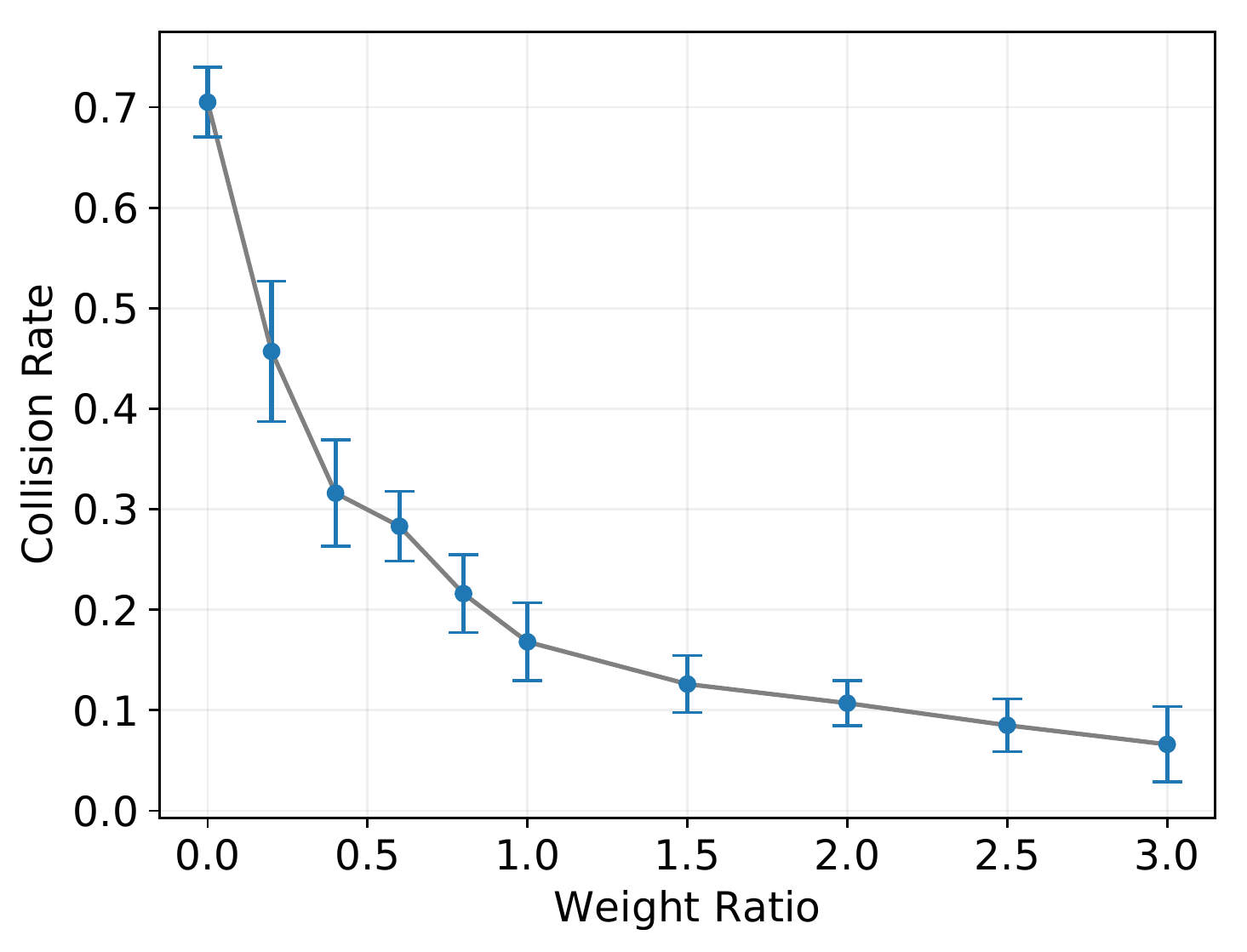} 
  \caption{Weight ratio versus collision rate.}
  \label{fig:ratio_collision}
\end{minipage}
\end{figure*}

\section{Experiments and Results}
In this section, we first introduce the scenario that we used in the experiment. Afterwards, we evaluate the prediction results with only the learning-based part of the proposed architecture. Finally, we assess the quality of the proposed model architecture in different aspects.
\subsection{Real-world Scenario}
We conduct experiments on a roundabout scenario included in the INTERACTION dataset \cite{Wei2019IROS, zhan_2019}. It is a 8-way roundabout and each of the branch has one entry lane and one exit lane. The bird-view image of the roundabout as well as the reference path information can be found in \cite{Wei2019IROS, Yeping_2019IV, zhan_2019}. 
We manually selected 1120 highly interactive driving segments in the dataset and used 80 \%  for training and the remaining for testing.  We define the car that is about to enter the circular roadway as the ego vehicle and the car that is already driving on the circular roadway as the predicted vehicle. In this problem setting, our goal is to predict 1s into the future using the past 1s information with a sampling frequency of 5Hz.

\subsection{Learning-based Trajectory Prediction}

\subsubsection{Prediction Accuracy}
We calculated the root mean squared error (RMSE) between the predicted and ground-truth state for both interacting vehicles at each future time step. The results are shown in the table below. According to the table, the mean RMSE error and the standard deviation continuously increase as the prediction horizon extends. However, prediction errors for both vehicles are all within an acceptable range even when the prediction horizon reaches one second. 


\begin{table}[ht]
	\caption{Evaluation Results (m)}
	\label{tab:pred_accuracy}
	\centering
	\begin{tabular}{p{0.4cm} p{1.15cm} p{1.15cm} p{1.15cm} p{1.15cm} p{1.15cm}}
		\toprule 
		 & 0.2s & 0.4s & 0.6s & 0.8s & 1.0s \\
		\midrule \midrule
		$ego$ & 0.07$\pm$0.01 & 0.16$\pm$0.05 & 0.29$\pm$0.12 & 0.40$\pm$0.20 & 0.51$\pm$0.29 \\
		\midrule
		$pred$ & 0.08$\pm$0.02 & 0.18$\pm$0.04 & 0.33$\pm$0.10 & 0.50$\pm$0.18 & 0.70$\pm$0.26 \\
		\bottomrule
	\end{tabular}
\end{table}
\subsubsection{Generalization Ability}
To illustrate that the proposed prediction algorithm is able to generalize well under unseen scenarios, we select a test data from another entrance of the roundabout that is not considered in the training data. As shown in Fig.~\ref{fig:generalization}, although the location of the interaction changes, our architecture can still have reasonable prediction results and can generate multi-modal trajectories. 
\subsubsection{Failures under Unseen/Corner Cases}
Even if the learning-based prediction method has been proved to have good testing performances in the previous section, the model can still fail under some corner cases that have certain discrepancies from the collected data as discussed in Section I-B. To demonstrate potential failure testing cases, we formulated four different artificial test scenarios that are unseen from the training set. We fixed $\xi^{pred}$ and the initial state of the ego vehicle and assumed that the ego vehicle would drive at constant speed during the historical time steps. We then assigned four different velocities to the ego vehicle: \{4$m/s$, 6$m/s$, 7$m/s$, 8$m/s$\}. 

From the prediction results in Fig.~\ref{fig:artificial_test}, the sampled trajectories in case (a) and (d) are all feasible and collision free. When ego vehicle's historical motion changes slightly as in (b) and (c), infeasible trajectories are predicted, where a collision occurs. However, we cannot simply conclude that the prediction method failed in every cases that a collision is predicted since under some dangerous circumstances, the predictor is expected to generate results that can reflect such situation. 

Therefore, to further analyze if (b) and (c) are indeed failure test cases, we first checked whether the two vehicles have any chances to avoid collision given their initial states for each testing case. We applied a constant deceleration model on one vehicle while letting the other vehicle drive with constant speed. Our result shows that if any of the two vehicles brakes in time, collision can be avoided. Therefore, the predicted result in all four scenarios should be expected to be either collision free or have small collision rate. However, as observed in Fig.~\ref{fig:artificial_test}, all the sampled joint trajectories in case (b) and more than half percent of the samples in case (c) are infeasible. Therefore, we concluded that that these two scenarios are indeed failure testing cases as they violate the common consensus of human behaviors, which verifies the drawback of using pure learning-based method for trajectory prediction.


\subsection{Overall Framework Evaluation} 
\subsubsection{Convert to Conditional Distribution}
In order to convert the predicted joint distribution to conditional distribution, we need to filter out the predicted trajectories for the ego vehicle $\hat{\xi}^{ego}$ that are far from its planned ground-truth trajectory $\hat{\xi}^{ego}_{gt}$. 

We obtained the trajectory discrepancies by calculating the RMSE error between the last state of each predicted trajectory and that of the actual trajectory. We set the discrepancy threshold to $0.2$ meter and thus the sampled joint trajectory will be retained only when $|\hat{\xi}^{ego}_{gt}(t_{final}) - \hat{\xi}^{ego}(t_{final})| \leq 0.2 m$, where $t_{final}$ is the the final prediction time step. 

\begin{figure}[htbp]
	\centering
	\includegraphics[scale=0.25]{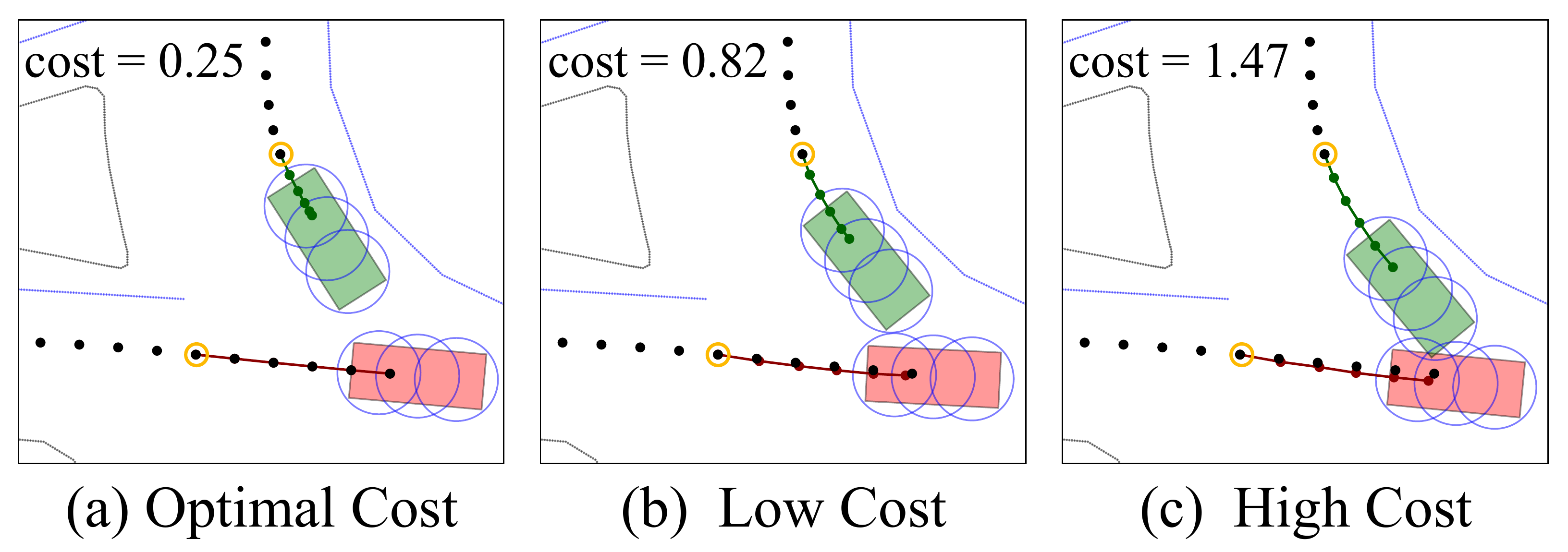}
	\caption{Different sampled trajectories and their corresponding costs at a selected scenario.}
	\label{fig:different_feasibility}
\end{figure}

\subsubsection{Trajectory Cost Evaluation}
We defined three categories of driving behaviors for the predicted vehicle: optimal, rational, and irrational. In order to demonstrate the cost differences among these categories, we plotted one sampled trajectory from each category and calculated the corresponding cost $C$ after acquiring the parameters $\bm{\theta}$ of the cost function from the IRL algorithm (Fig.~\ref{fig:different_feasibility}). According to the corresponding cost of each sample, we can conclude that the learned IRL cost parameters are able to assign high cost when collision occurs due to irrational behavior (Fig.~\ref{fig:different_feasibility}(c)) and low cost when the predicted trajectory results in rational behavior (Fig.~\ref{fig:different_feasibility}(b)). In this testing scenario, the optimal cost is achieved when the predicted vehicle applies maximum deceleration (4$m/s^2$) at every future time steps as shown in Fig.~\ref{fig:different_feasibility}(a). 

\subsubsection{Distribution Re-weighting}
One of the most important aspects needs to be evaluated is whether we are indeed able to reasonably change the original sample distribution. Besides, we need to further examine the correlations between the weight ratio factor $r$ and the final outcomes. First of all, under a selected testing scenario, we let $N_{opt} = 1$ during the 4$^{th}$ step in Algorithm 1, where we added only one optimal trajectory to the current sample set and resampled $N$ final trajectories for evaluation. We then gradually increased $N_{opt}$ while keep sampling the same number of final samples. Lastly, we calculated the collision rate of the sampled trajectory set corresponds to each $r$ and plotted the result in Fig.~\ref{fig:ratio_collision}. 

From the plot, we notice that when $r = 0$, where no optimal trajectory is added, the collision rate is 0.7 from the pure learning-based prediction results. As the planning-based model is introduced to the algorithm and the weight ratio gradually increases, the collision rate decreases dramatically at the beginning and slowly converges to 0 as $r$ passes 1. Such result is reasonable since when more rational and feasible samples with lower costs are added to the sample set, the original conditional distribution will not only gradually shifts away from the infeasible sample area horizontally but also gets lower probability at those irrational sample points as illustrated in the reweighting part of Fig.~\ref{fig:architecture}. Therefore, we concluded that the proposed architecture is capable of reshaping the original sample distribution to desired outcomes when it is used online with a constantly updating $r$ value. 

\begin{figure}[htbp]
	\centering
	\includegraphics[scale=0.31]{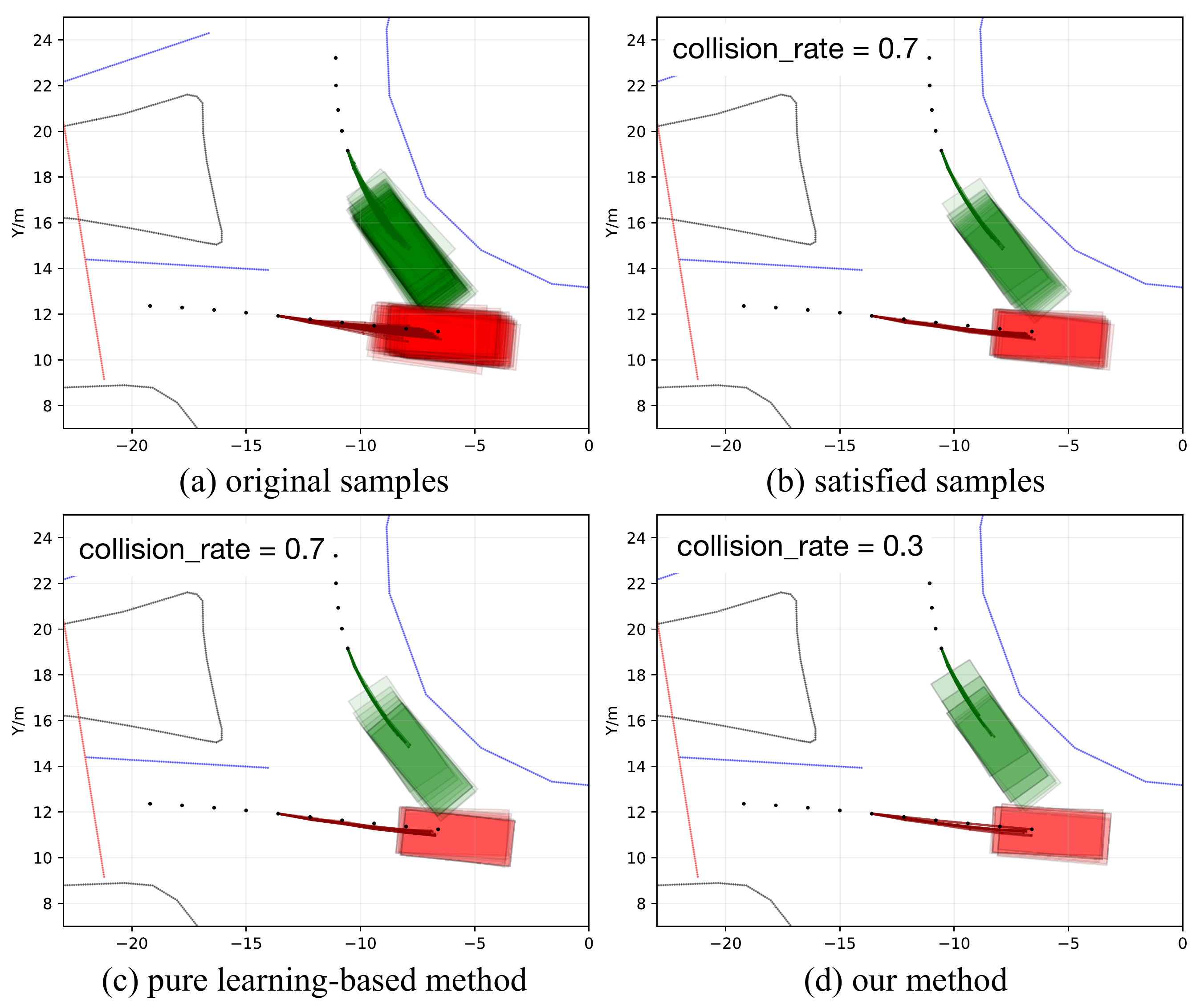}
	\caption{Visualization of the overall framework at different stages and a result comparison with the pure learning-based prediction method.}
	\label{fig:result_visualization}
\end{figure}

\subsubsection{Final Results Visualization}
We selected the same scenario as in 3) to compare the prediction result of our framework with the pure learning-based prediction method. Note that the online path planning for ego vehicle is not the focus of our current work and thus we will not visualize the prediction result of applying online update to the weight ratio. Instead, we assigned $r$ to a fixed ratio and visualized the prediction result at a single time step (Fig.~\ref{fig:result_visualization}).

The result of directly sampling from the learned learning-based prediction models is shown in Fig.~\ref{fig:result_visualization}(a), which corresponds to step 1) in Algorithm 1. After converting the joint distribution to conditional distribution in step 2), the remaining satisfied trajectories are shown in Fig.~\ref{fig:result_visualization}(b), where we notice that all the ego vehicle's predicted trajectories are close to the ground-truth as desired. The collision rate in (b) is 0.7, which implies that more irrational than rational behaviors are predicted. We then directly sampled from the current sample set and obtained the prediction results by using only the learning-based method as in Fig.~\ref{fig:result_visualization}(c). Comparing to the predicted results in Fig.~\ref{fig:result_visualization}(c) , the generated trajectories of our approach in Fig.~\ref{fig:result_visualization}(d) tend to have more rational behaviors while still having a collision rate of 0.3 as a warning to the ego vehicle. In this way, the ego vehicle is able to have both rational and irrational information about the predicted vehicle's future behavior, where it believes that the predicted vehicle will be more likely to behave rationally in the future while preparing for possible irrational behaviors.


\section{Conclusions}
In this paper, a generic prediction architecture is proposed, which can predict continuous trajectories of other vehicles by considering both rational and irrational driving behaviors. An exemplar roundabout scenario with real-world data was used to demonstrate the performance of our method. We first evaluated the prediction accuracy of the learning-based method on the test dataset. Then, we demonstrated the generalizability of the method by using our generic environmental representation. By testing on some unseen and corner driving scenarios, we revealed the limitations of using pure learning-based prediction method. Finally, by thoroughly examining the proposed architecture, we concluded that the approach of combining both learning-based and planning-based method can enhance the overall prediction performance by providing sufficient possible outcomes to the ego vehicle. For future work, we will perform human-in-the-loop experiments online using the proposed prediction architecture to evaluate its capabilities. 

\addtolength{\textheight}{-5cm}   


\bibliographystyle{IEEEtran}
\bibliography{ITSC2019_ref}

\end{document}